\documentclass[10pt,twocolumn,letterpaper]{article}

\usepackage{iccv}
\usepackage{times}
\usepackage{epsfig}
\usepackage{graphicx}
\usepackage{amsmath}
\usepackage{amssymb}

\usepackage[breaklinks=true,bookmarks=false]{hyperref}
\usepackage{array}
\usepackage{multirow}
\iccvfinalcopy 

\def\iccvPaperID{****} 

\makeatletter
\newcommand{\printfnsymbol}[1]{%
  \textsuperscript{\@fnsymbol{#1}}%
}
\makeatother
\begin{document}

\title{2D Attentional Irregular Scene Text Recognizer}

\author{Pengyuan Lyu\thanks{Authors contribute equally.} \textsuperscript{1}, Zhicheng Yang\printfnsymbol{1}\textsuperscript{1,2}, Xinhang Leng\textsuperscript{1,3}, Xiaojun Wu\textsuperscript{2}, Ruiyu Li\textsuperscript{1},  Xiaoyong Shen\textsuperscript{1}\\
\textsuperscript{1}YouTu Lab, Tencent
\textsuperscript{2}Harbin Institute of Technology
\textsuperscript{3}The Chinese University of Hong Kong\\
{\tt\small \{pengyuanlv, cosinyang, xinhangleng, royryli, dylanshen\}@tencent.com, wuxj@hit.edu.cn}}

\maketitle

\begin{abstract}
   Irregular scene text, which has complex layout in 2D space, is challenging to most previous scene text recognizers. Recently, some irregular scene text recognizers either rectify the irregular text to regular text image with approximate 1D layout or transform the 2D image feature map to 1D feature sequence. Though these methods have achieved good performance, the robustness and accuracy are still limited due to the loss of spatial information in the process of 2D to 1D transformation.  Different from all of previous, we in this paper propose a framework which transforms the irregular text with 2D layout to character sequence directly via 2D attentional scheme. We utilize a relation attention module to capture the dependencies of feature maps and a parallel attention module to decode all characters in parallel, which make our method more effective and efficient.  Extensive experiments on several public benchmarks as well as our collected multi-line text dataset show that our approach is effective to recognize regular and irregular scene text and outperforms previous methods both in accuracy and speed.
\end{abstract}

\section{Introduction}


Recently, reading text from image especially in natural scene has attracted much attention from the academia and industry, due to the huge demand in a large number of applications. Text recognition, as the indispensable part, plays a significant role in an OCR system~\cite{DBLP:conf/iccv/WangBB11,DBLP:conf/cvpr/NeumannM12,DBLP:conf/icpr/WangWCN12,DBLP:journals/tip/YaoBL14,DBLP:journals/ijcv/JaderbergSVZ16,DBLP:conf/iccv/BustaNM17,DBLP:conf/cvpr/HeTHS0S18,DBLP:conf/cvpr/LiuLYCQY18,DBLP:conf/eccv/LyuLYWB18}. 

Despite large amount of approaches~\cite{DBLP:conf/iccv/WangBB11,DBLP:conf/cvpr/NeumannM12,DBLP:conf/icpr/WangWCN12,DBLP:journals/tip/YaoBL14,DBLP:conf/cvpr/YaoBSL14,DBLP:conf/eccv/JaderbergVZ14,DBLP:conf/cvpr/LeeBDJP14,DBLP:journals/corr/JaderbergSVZ14, DBLP:journals/ijcv/JaderbergSVZ16,DBLP:journals/pami/ShiBY17,ShiWLYB16,DBLP:conf/cvpr/LeeO16} were proposed, scene text recognition is still challenging. Except for the complex background and the varied appearance, the irregular layout further increases the difficulty. Moreover, since most  previous methods are designed for the recognition of regular text with approximate 1D layout, they do not have the ability to handle the irregular text image (such as curved  or multi-line text image) that  the characters are distributed in 2D space.

\begin{figure*}
\begin{centering}
\includegraphics[scale=1.1]{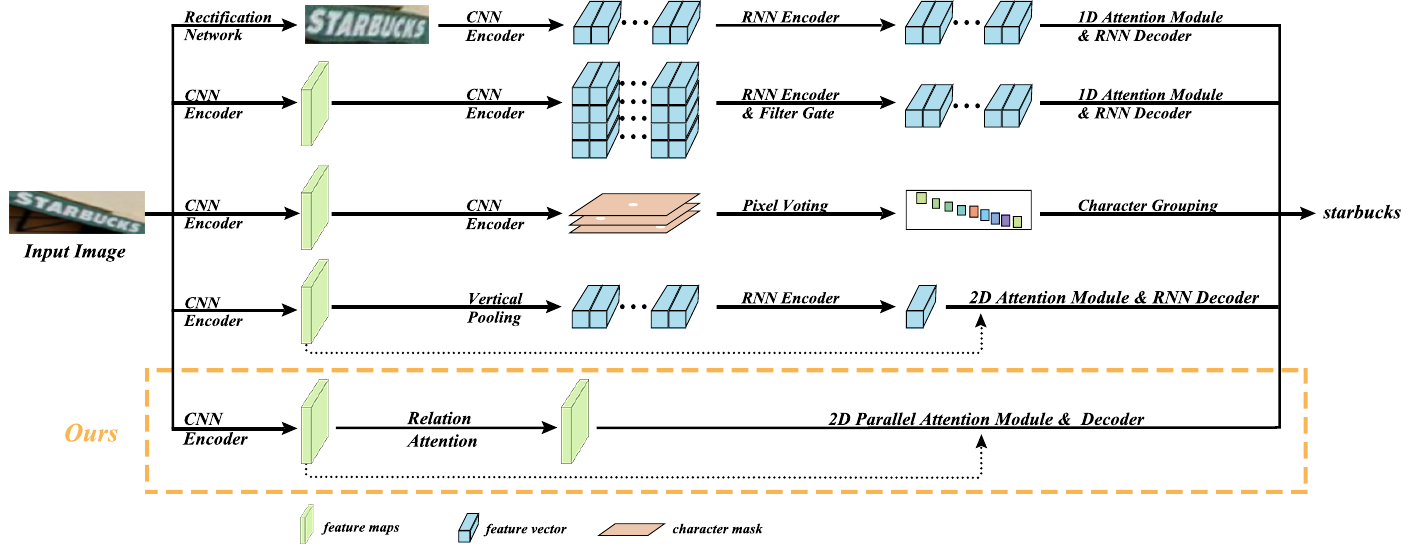}
\par\end{centering}
\caption{Comparison of several recent irregular scene text recognition works' pipelines. The first branch: the pipeline of~\cite{ShiWLYB16,bshi2018aster,cluo2019moran}. The second branch: the pipeline of~\cite{DBLP:conf/cvpr/ChengXBNPZ18}. The third branch: the pipeline proposed by~\cite{DBLP:conf/eccv/LyuLYWB18,DBLP:journals/corr/abs-1809-06508}. The fourth branch: the pipeline proposed by~\cite{DBLP:journals/corr/abs-1811-00751}. The last branch: our proposed pipeline.}
\label{relatedwork}
\vspace{-5mm}
\end{figure*}

\vspace{-0.1in}
\paragraph{Status of Current Irregular Text Recognizer}
Many methods are proposed to recognize irregular text image these years. We show the representative pipelines in Figure~\ref{relatedwork}. 
Straight-forward methods are proposed in ~\cite{ShiWLYB16,bshi2018aster,cluo2019moran} which first rectify the irregular text image into regular image, and then recognize the rectified one with normal text recognizer. To address the irregular information, 
Cheng~\etal ~\cite{DBLP:conf/cvpr/ChengXBNPZ18} encode 2D space information from four directions to transform the 2D image feature maps to 1D feature sequences
while Lyu~\etal \cite{DBLP:conf/eccv/LyuLYWB18} and Liao~\etal \cite{DBLP:journals/corr/abs-1809-06508} apply semantic segmentation to yield 2D character mask, and then group the characters with the character position and heuristic algorithm as shown in the second and third branch respectively in Figure~\ref{relatedwork}.  Recently, Li~\etal~\cite{DBLP:journals/corr/abs-1811-00751} propose an encoder-decoder based scheme as illustrated in the fourth branch in Figure~\ref{relatedwork}, where an encoder extracts holistic feature via encoding the 2D feature map column by column and a decoder decodes the holistic feature to character sequence recurrently by applying 2D attention on image feature maps. 

Though promising results are achieved by the mentioned representative approaches, there are still some limitations preventing these methods to be more flexible, robust and efficient. First, methods~\cite{DBLP:conf/ijcai/YangHZKG17,DBLP:conf/eccv/LyuLYWB18,DBLP:journals/corr/abs-1809-06508}  need character-level localization in the training phase. It is usually limited when no character-level annotations are provided. Second, many schemes are not robust enough to  handle very complex cases such as large curved text or multi-line text, due to: 1) the insufficient capacity of the rectification network~\cite{ShiWLYB16,bshi2018aster,cluo2019moran}; 2) the limitation of the character grouping algorithm~\cite{DBLP:conf/eccv/LyuLYWB18, DBLP:journals/corr/abs-1809-06508}; 3) the loss of spatial information~\cite{ShiWLYB16,bshi2018aster,cluo2019moran,DBLP:conf/cvpr/ChengXBNPZ18,DBLP:journals/corr/abs-1811-00751,DBLP:conf/eccv/LyuLYWB18, DBLP:journals/corr/abs-1809-06508}. Third, some approaches~\cite{ShiWLYB16,bshi2018aster,cluo2019moran,DBLP:conf/aaai/LiuCW18,DBLP:conf/ijcai/YangHZKG17,DBLP:journals/corr/abs-1811-00751}  in encoder-decoder framework always use RNN and attention module in serial, which is inefficient in sequential computation, especially when predicting long text.

\vspace{-0.1in}
\paragraph{Our Contributions}
To mitigate the limitations of previous, we propose a novel 2D attentional irregular scene text recognizer shown in the last branch in Figure~\ref{relatedwork}. Different from all of previous, ours directly encode and decode text information in 2D space through all the pipeline by 2D attentional modules. 
To achieve the goal, we first propose the relation attention module to capture the global context information instead of modeling the context information with RNN as in~\cite{DBLP:conf/cvpr/ChengXBNPZ18, DBLP:conf/ijcai/YangHZKG17, DBLP:journals/corr/abs-1811-00751}. 
In addition, we design a parallel attention module which yields multiple attention masks on 2D feature map in parallel. With this module, our method can output all the characters at the same time, rather than predict the characters one by one as previous methods~\cite{ShiWLYB16,bshi2018aster,DBLP:conf/aaai/LiuCW18,DBLP:conf/ijcai/YangHZKG17,DBLP:journals/corr/abs-1811-00751}. 
Contrary to previous, ours is the end-to-end trainable framework without complex post-processing to detect and group characters and character-level or pixel-level annotations for training. With the relation module, our framework models the local and global context information which is more robust to handle complex irregular text (such as large curved or multi-line text). Our method is also more efficient since the proposed parallel module simultaneously predicts all results compared with previous RNN schemes. 

To verify the effectiveness, we conduct experiments on 7 public benchmarks that contain both regular and irregular datasets. We achieved state-of-the-art results on almost all datasets which demonstrate the advantages of the proposed algorithm. Especially, on SVTP and CUTE80, our method beats the previous best method~\cite{bshi2018aster} and~\cite{DBLP:journals/corr/abs-1811-00751} by $3.8\%$ and $3.5\%$, respectively. Besides, to evaluate the capability of our method on complex scenarios, we collect a cropped license plate image dataset which contains  text in one-line and multi-line. On this dataset, our method outperforms the rectification-based method ~\cite{bshi2018aster} and the recurrent 2D attention-based method~\cite{DBLP:journals/corr/abs-1811-00751} by $18.2\%$ and $29.6\%$ respectively, which proves the robustness of our framework. Moreover, our approach is \textbf{2.1} and \textbf{4.4} times faster than~\cite{bshi2018aster} and~\cite{DBLP:journals/corr/abs-1811-00751}, respectively.

In summary, the major contribution of this paper is three-fold:
1) An effective and efficient irregular scene text recognizer were proposed which is designed with 2D attentional  modules and achieved state-of-the-art results both on regular and irregular scene text datasets
2) We proposed 2D relation attention module and parallel attention module making the framework more flexible, robust and efficient; 3) A new dataset containing text in multi-line is constructed. As far as we know, our method is the first to show the capacity of recognizing cropped scene text in multi-line.

\begin{figure*}
\begin{centering}
\includegraphics[scale=1.0]{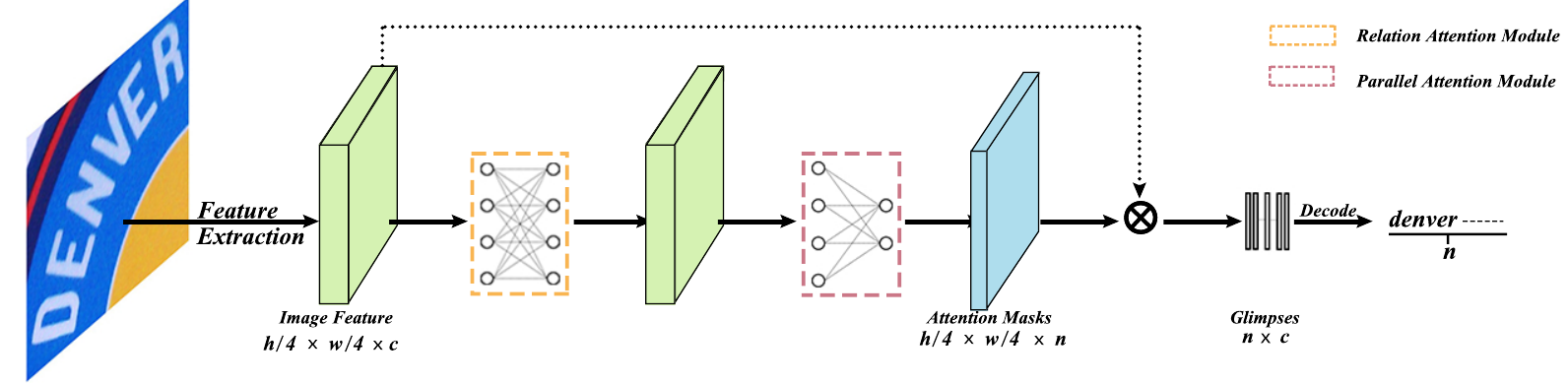}
\par\end{centering}
\caption{Overview of the proposed method. ``$c$'' means the number of channels of the extracted image feature. ``$h$" and ``$w$" are the height and width of the input image. ``$n$" represents the number of output nodes. ``-" is a special symbol which means ``End of Sequence"(EOS).}
\label{framework}
\vspace{-5mm}
\end{figure*}

\section{Related Work}

\subsection{Scene Text Recognition}
In recent years, a large number of methods have been proposed to recognize text in natural scene. Based on the characteristics of these methods, most of the methods fall into the following three categories: character-based methods, word-based methods and sequence-based methods. 

The early works~\cite{DBLP:conf/iccv/WangBB11,DBLP:conf/cvpr/NeumannM12,DBLP:conf/icpr/WangWCN12,DBLP:journals/tip/YaoBL14,DBLP:conf/cvpr/YaoBSL14,DBLP:conf/eccv/JaderbergVZ14,DBLP:conf/cvpr/LeeBDJP14} are mostly character-based. They first detect and classify individual characters, and then group them into words.  In most cases, the characters proposals are yielded  via connected components~\cite{DBLP:conf/cvpr/NeumannM12, DBLP:journals/tip/YaoBL14} or sliding window~\cite{DBLP:conf/iccv/WangBB11,DBLP:conf/cvpr/YaoBSL14,DBLP:conf/icpr/WangWCN12,DBLP:conf/eccv/JaderbergVZ14,DBLP:conf/cvpr/LeeBDJP14}, and then classified with hand-crafted feature (\emph{e.g.}, HOG~\cite{DBLP:conf/cvpr/YaoBSL14}) or learned feature~\cite{DBLP:conf/icpr/WangWCN12,DBLP:conf/eccv/JaderbergVZ14}. Finally, the individual characters are merged into a word with a heuristic algorithm.

Recently, some word-based methods~\cite{DBLP:journals/corr/JaderbergSVZ14, DBLP:journals/ijcv/JaderbergSVZ16} and sequence-based methods~\cite{DBLP:journals/pami/ShiBY17,ShiWLYB16,DBLP:conf/cvpr/LeeO16}  equipped with CNN~\cite{lecun1998gradient} and RNN~\cite{DBLP:journals/neco/HochreiterS97, DBLP:conf/nips/ChorowskiBSCB15} are proposed with the development of deep learning. In~\cite{DBLP:journals/corr/JaderbergSVZ14, DBLP:journals/ijcv/JaderbergSVZ16}, Jaderberg~\emph{et al.} solve the problem of text recognition with multi-class classification.
Based on 8 million synthetic images, they train a powerful classifier, which classifies the 90K common English  words directly. In~\cite{DBLP:journals/pami/ShiBY17,ShiWLYB16,DBLP:conf/cvpr/LeeO16}, the text images are recognized in a sequence-to-sequence manner. Shi~\emph{et al.}~\cite{DBLP:journals/pami/ShiBY17} first transfer the input image into a  feature sequence with CNN and RNN and obtain the recognition consequence with CTC~\cite{DBLP:conf/icml/GravesFGS06}. In~\cite{ShiWLYB16,DBLP:conf/cvpr/LeeO16}, the sequence is generated by the attention model in~\cite{DBLP:journals/corr/BahdanauCB14,DBLP:conf/nips/ChorowskiBSCB15}, step by step.

\subsection{Irregular Scene Text Recognition}
Irregular scene text recognition has also attracted much attention in recent years. In~\cite{ShiWLYB16,bshi2018aster,cluo2019moran}, Shi~\emph{et al.} and Luo~\emph{et al.} propose a unified network which contains a rectification network and a recognition network to recognize irregular scene text. They first use a Spatial Transform Network~\cite{DBLP:conf/nips/JaderbergSZK15} to transfer the irregular input image into a regular image which is easy to be recognized by the recognition network, and then recognize the transformed image with a sequence-based recognizer. Instead of the methods in ~\cite{ShiWLYB16,bshi2018aster,cluo2019moran} which rectify the whole input image directly, Liu~\emph{et al.}~\cite{DBLP:conf/aaai/LiuCW18} propose to rectify the individual characters recurrently. In~\cite{DBLP:conf/cvpr/ChengXBNPZ18, DBLP:conf/ijcai/YangHZKG17, DBLP:journals/corr/abs-1811-00751,DBLP:conf/eccv/LyuLYWB18, DBLP:journals/corr/abs-1809-06508}, the authors propose to recognize irregular text images in 2D perspective. In~\cite{DBLP:conf/cvpr/ChengXBNPZ18}, to recognize irregular text image with arbitrarily-oriented, Cheng~\emph{et al.} adapt the sequence-based model~\cite{ShiWLYB16,DBLP:conf/cvpr/LeeO16} with the image feature extracted from four directions. In~\cite{DBLP:conf/ijcai/YangHZKG17,DBLP:journals/corr/abs-1811-00751}, the irregular images are handled
via applying 2D attention mechanism on feature maps. In~\cite{DBLP:conf/eccv/LyuLYWB18, DBLP:journals/corr/abs-1809-06508}, with the character-level localization annotations, Lyu~\emph{et al.} and Liao~\emph{et al.} recognize irregular text via pixel-wise semantic segmentation.

In this paper, we propose a 2D attentional irregular scene text recognizer which transforms the irregular text with 2D
layout to character sequence directly. Compared to the rectification pipeline~\cite{ShiWLYB16,bshi2018aster,DBLP:conf/aaai/LiuCW18,cluo2019moran}, our approach is more robust and effective to recognize irregular scene text image. Besides, compared to the previous methods~\cite{DBLP:conf/cvpr/ChengXBNPZ18, DBLP:conf/ijcai/YangHZKG17, DBLP:journals/corr/abs-1811-00751,DBLP:conf/eccv/LyuLYWB18, DBLP:journals/corr/abs-1809-06508} in 2D perspective, our method has parallel attention mechanism, which is more efficient than~\cite{DBLP:conf/ijcai/YangHZKG17,DBLP:journals/corr/abs-1811-00751} predicting sequence serially, and doesn't need character-level localization annotations as~\cite{DBLP:conf/ijcai/YangHZKG17,DBLP:conf/eccv/LyuLYWB18, DBLP:journals/corr/abs-1809-06508}.

\section{Methodology}
The proposed model is a unified network which can be trained and evaluated end-to-end. Given an input image, the network predicts the recognition consequence directly.

\subsection{Network Structure}

The network structure is shown in Figure~\ref{framework}. Given an input image, we first use a CNN encoder to transform the input image into feature maps with high-level semantic information. Then, a relation attention module is applied to each pixel of the feature maps to capture global dependencies. After that, the parallel attention module is built on the output of relation attention module and outputs a fixed number of glimpses. Finally, the character decoder decodes the  glimpses into characters.

\subsubsection{Relation Attention Module}
Previous methods such as~\cite{DBLP:journals/pami/ShiBY17,ShiWLYB16,bshi2018aster} always use RNN to capture the dependencies of the CNN encoded 1D feature sequence, but it is not a good choice to apply RNN on 2D feature maps directly for the consideration of computational efficiency. In~\cite{DBLP:conf/cvpr/ChengXBNPZ18}, Cheng~\emph{et al.} apply RNN on four 1D feature sequences that extracted from four directions of the 2D feature maps. In~\cite{DBLP:journals/corr/abs-1811-00751}, Li~\emph{et al.} convert the 2D feature maps into 1D feature sequence with a max-pooling along the vertical
axis. The strategies used in~\cite{DBLP:conf/cvpr/ChengXBNPZ18,DBLP:journals/corr/abs-1811-00751} can reduce  computation to some extent, but meanwhile, some space information may also be lost. 

Inspired by~\cite{DBLP:conf/nips/VaswaniSPUJGKP17,DBLP:conf/cvpr/HuGZDW18,DBLP:journals/corr/abs-1810-04805} which capture the global dependencies between input and output by aggregating information from the elements of the input, we build a relation attention module which consists of the~\emph{transformer} unit proposed in~\cite{DBLP:conf/nips/VaswaniSPUJGKP17}. The relation attention module captures the global information in parallel, which is more efficient than the above-mentioned strategies. Specifically, following BERT~\cite{DBLP:journals/corr/abs-1810-04805}, the architecture of our relation attention module is a multi-layer bidirectional transformer encoder. We present it in the supplementary material due to the page limit.

To apply the relation attention module to the input with arbitrary shape conveniently, we flat the input feature maps into a feature sequence $I$ with the shape of $k \times c$. The $k$ means the length of the flatted feature sequence, and $c$ is the dimensions of each feature vector in the feature sequence. For each feature vector $I_{i}(i\in[1, k])$, we use an embedding operator to encode the position index $i$ into position vector $E_{i}$, which has the same dimensions as $I_{i}$. After that, the flatted input feature $I$ and the position-embedded feature $E$ are added to form the fused feature $F$ which is sensitive to the position.




We build several transformer layers in series to aggregate information from the fused feature. Each transformer layer consists of $k$ transformer units. And for each transformer,   the~\emph{query},~\emph{keys} and~\emph{values}~\cite{DBLP:conf/nips/VaswaniSPUJGKP17} are obtained with the following process:
\begin{equation} 
Q_{l}^{i} = \left\{  
             \begin{aligned} 
              &F_{i} \quad  &l = 1 \\  
              &O_{l-1}^{i} \quad &l > 1
             \end{aligned}  
\right.,
\end{equation}

\begin{equation}
K_{l}^{i} = \left\{  
             \begin{aligned} 
              &F \quad  &l = 1 \\  
              &O_{l-1} \quad &l > 1
             \end{aligned}  
\right.,
\end{equation}

\begin{equation}
V_{l}^{i} = \left\{  
             \begin{aligned} 
              &F \quad  &l = 1 \\  
              &O_{l-1} \quad &l > 1
             \end{aligned}  
\right..
\end{equation}
Here, $Q_{l}^{i}$ is the query vector of $i$-th transformer unit with the shape of $1 \times c$ in $l$-th transformer layer; $K_{l}^{i}$ and $V_{l}^{i}$ are the key vectors and the value vectors, both of which are in the shape of $k \times c$; $O_{l-1}$ means the output of the previous transformer layer which is also in the shape of $k \times c$.

With the query, keys and the values as input, the output of a transformer is computed by a weighted sum operator  applied to the values. And the weights of each value is calculated as the following formulas:

\begin{equation}
    \alpha_{l}^{ij} = \frac{exp(W_{l}^{q}Q_{l}^{i} \cdot W_{l}^{k}K_{l}^{ij})}{\sum_{j^{'}=1}^{k} exp(W_{l}^{q}Q_{l}^{i} \cdot W_{l}^{k}K_{l}^{ij^{'}})},
\end{equation} where $W$ are trainable weights.

Taking the weights $\alpha$ as the coefficients, the output of each transformer is weighted and summed as:
\begin{equation}
    O_{l}^{i} = Func( \sum_{j=1}^{k}\alpha_{l}^{ij}  W_{l}^{v}V_{l}^{ij}),
\end{equation} where $W_{l}^{v}$ is a learned weight. $Func$ is a non-linear function, which can be referred to~\cite{DBLP:conf/nips/VaswaniSPUJGKP17,DBLP:journals/corr/abs-1810-04805} for detail.

We take the outputs of the last transform layer as the relation attention module's output.  

\subsubsection{Parallel Attention Module}
The basic attention module used in~\cite{ShiWLYB16,bshi2018aster,DBLP:conf/ijcai/YangHZKG17,DBLP:journals/corr/abs-1811-00751}  always work in serial and be integrated with a RNN:
\begin{equation}
    \alpha_{t} = Attention(h_{t-1}, \alpha_{t-1}, I),
    \label{recurrent attention}
\end{equation} where $h_{t-1}$ and $\alpha_{t-1}$ are the hidden state  and attention weights of the RNN decoder at the previous step, $I$ means the encoded image feature sequence. As formulated in Eq.~\ref{recurrent attention}, the computation of the step $t$ is limited by the previous steps, which is inefficient.

Instead of attending recurrently, we propose a parallel attention module, in which the dependency relationships of the output nodes are removed. Thus, for each output node, the computation of attention is independent, which can be implemented and optimized in parallel easily.

Specifically, we assign the number of output nodes to $n$. Given a feature sequence $O$ in the shape of $k \times c$, the parallel attention module outputs the weight coefficient $\alpha$ with the following process:
\begin{equation}
   \alpha = softmax(W_{2}tanh(W_{1}O^{T})).
\end{equation} Here, $W_{1}$ and $W_{2}$ are the learnable parameters with the shape of $c\times c$ and $n\times c$ respectively.

Based on the weight coefficients $\alpha$ and the encoded image feature sequence $I$, the glimpses of each output node can be obtained by:

\begin{equation}
   G_{i} = \sum_{j=1}^{k} \alpha_{ij}I_{j},
\end{equation}
where $i$ and $j$ are the index of output node and feature vector respectively.
\begin{figure}
\begin{centering}
\includegraphics[scale=0.35]{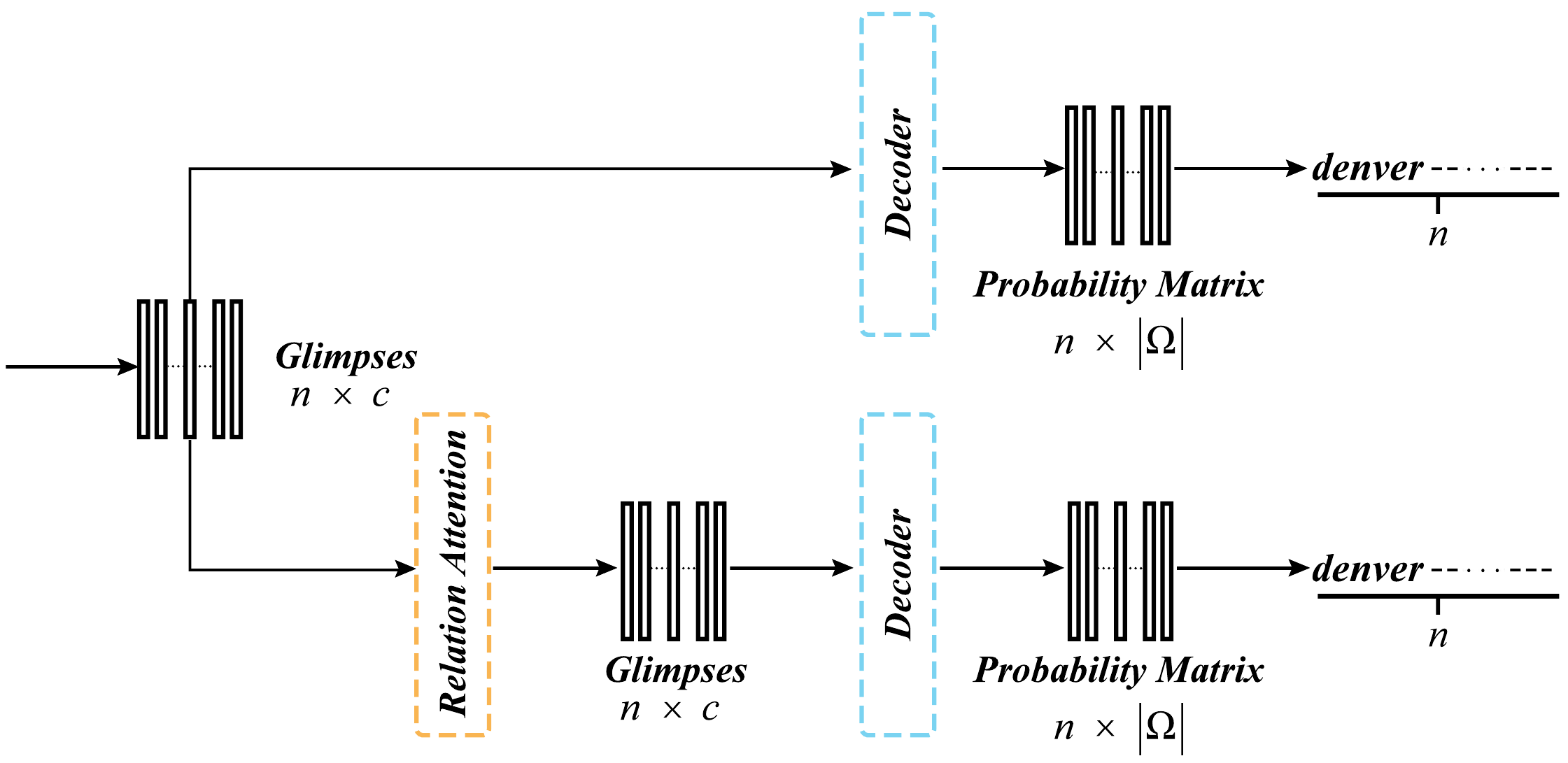}
\par\end{centering}
\caption{Overview of the two-stage decoder. ``$\Omega$" is the set of decoded characters, ``n" means the number of output nodes. ``c" is the number of channels of the extracted image feature maps.}
\label{decoder}
\vspace{-5mm}
\end{figure}



\subsubsection{Two-stage Decoder}

We take the glimpses to predict the characters with a character decoder. For each output node, the output character probability is predicted by:
\begin{equation}
   P_{i} = softmax(WG_{i} + b),
\end{equation} where $W$ and $b$ are the learned weights and bias.

Though the proposed parallel attention module is more efficient than the basic recurrent attention model, the dependency relationships among the output nodes are  lost due to the parallel independent computing. To capture the dependency relationships of output nodes, we build the second-stage decoder. The second-stage decoder consists of a relation attention module and a character decoder. As shown in Figure~\ref{decoder}, we take the glimpses as the input of relation attention module to model the dependency relationships of the output nodes. Besides, a character decoder is stacked over the relation attention module to produce the prediction results.

\subsection{Optimization}
We optimize the network in an end-to-end manner. The two decoders are optimized simultaneously with a multi-task loss:
\begin{equation}
   L = \sum_{i=1}^{2}\sum_{j=1}^{n}(-log P_{ij} (y_{j})),
\end{equation} where $y$ is the ground truth text sequence, $i$ and $j$ are the indexes of the decoder and the output node. In the training stage, a ground truth sequence will be padded with  symbol ``EOS" if the length of this sequence is shorter than $n$. By contrast, the excess parts will be discarded.

\section{Experiments}
\subsection{Datasets}
We conduct experiments on a number of datasets to verify the effectiveness of our proposed model. The model is trained on two  synthetic datasets: Synth90K~\cite{Synth90K} and SynthText~\cite{SynthText}, and evaluated on both regular and irregular scene text datasets. In particular, a license plate dataset which contains text in  one-line and multi-line is collected and proposed, for evaluating the capability of our model to recognize text with complex layout.

\textbf{Synth90K}~\cite{Synth90K} is a synthetic text dataset proposed by Jaderberg~\emph{et al.} in ~\cite{DBLP:journals/corr/JaderbergSVZ14}. This dataset is generated by randomly blending the 90K common English words on scene image patches. There are about 9 million images in this dataset and all of them are used to pre-train our model. 

\textbf{SynthText}~\cite{SynthText} is also a commonly used synthetic text dataset proposed in~\cite{DBLP:conf/cvpr/GuptaVZ16} by Gupta~\emph{et al.}. SynthText is generated for text detection and the text are randomly blended on full images rather than the image patches in~\cite{DBLP:journals/corr/JaderbergSVZ14}. All samples in the SynthText are cropped and token as the training data.

\textbf{ICDAR 2003 (IC03)}~\cite{DBLP:journals/ijdar/LucasPSTWYANOYMZOWJTWL05} is a real regular text dataset cropped from scene text images. After  filtering out some samples which contain non-alphanumeric characters or have less than three
characters as~\cite{DBLP:conf/iccv/WangBB11}, the dataset contains 860 images for test. Besides, for each sample, a 50-word lexicon is provided.

\textbf{ICDAR 2013 (IC13)}~\cite{DBLP:conf/icdar/KaratzasSUIBMMMAH13} is derived from IC03 with some new samples added. Following the previous methods~\cite{DBLP:journals/pami/ShiBY17,bshi2018aster}, we filter out some samples which contain non-alphanumeric characters, and keep the remaining 1015 images as the test data. In this dataset, no lexicon is provided.

\textbf{IIIT5k-Words (IIIT5K)}~\cite{DBLP:conf/bmvc/MishraAJ12} contains 5000 images, of which 3000 images are used for test. Most  samples in this dataset are regular. For each sample in the test set, two lexicons which contain 50 words and 1000 words respectively are provided.

\textbf{Street View Text (SVT)}~\cite{DBLP:conf/iccv/WangBB11} comprises 647 test samples cropped from Google Street View images. In this dataset, each image has a 50-word lexicon.

\textbf{SVT-Perspective (SVTP)}~\cite{DBLP:conf/iccv/PhanSTT13} also originates from Google Street View images. Since the texts are shot from the side view, many of the cropped samples are perspective distorted. So this dataset is usually used for evaluating the performance of recognizing perspective text. There are 639 samples for test, and for each sample, a 50-word lexicon is given.

\textbf{ICDAR 2015 Incidental Text (IC15)}~\cite{DBLP:conf/icdar/KaratzasGNGBIMN15} is cropped from the images shot with a pair of Google Glass in incidental scene. 2077 samples are included in the test set, and most of them are multi-oriented. 

\textbf{CUTE80 (CUTE)}~\cite{DBLP:journals/eswa/RisnumawanSCT14} is a dataset proposed for curved text detection, and is then used by~\cite{ShiWLYB16} to evaluate the capacity of a model to recognize curved text. There are 288 cropped patches for test and no lexicon is provided.

\begin{figure}
\begin{centering}
\includegraphics[scale=0.65]{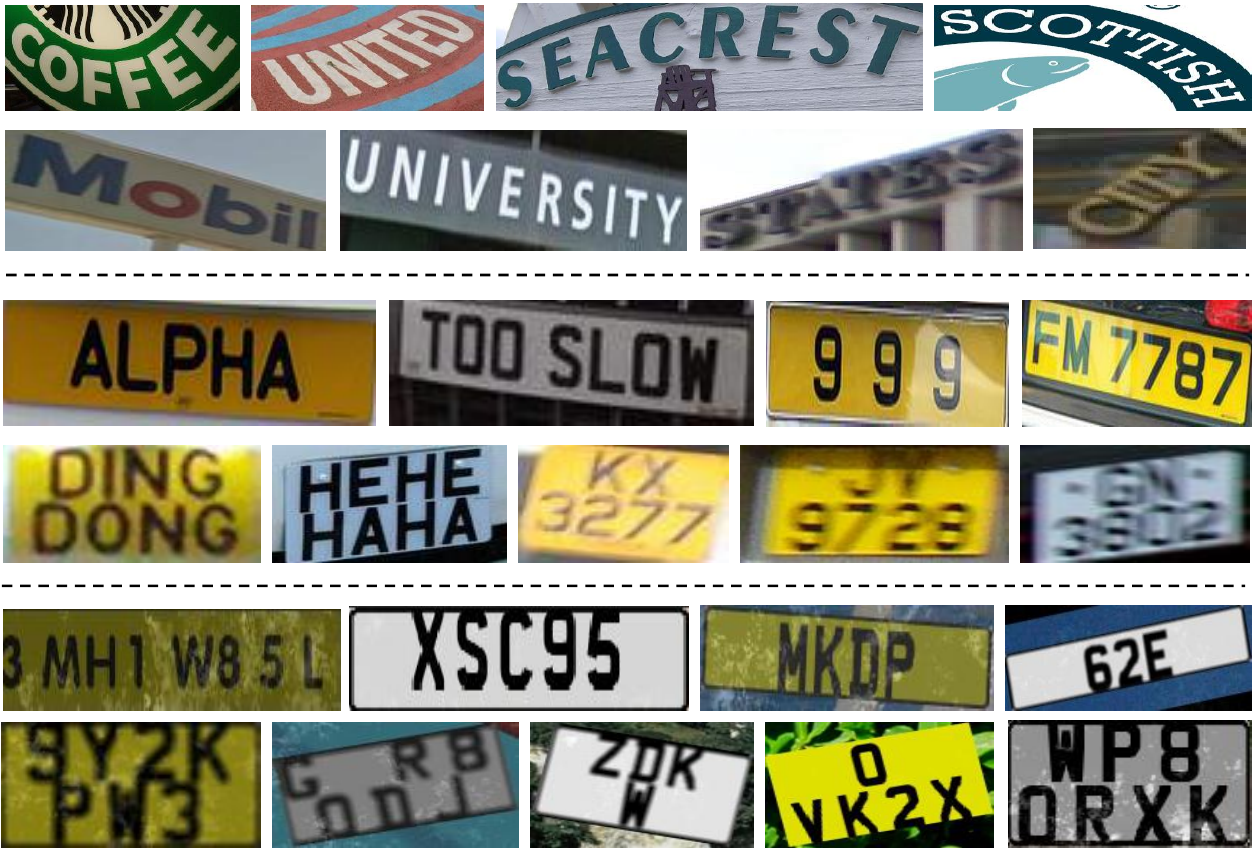}
\par\end{centering}
\caption{Visualization of some irregular scene text samples. The first two rows are samples from the above-mentioned 7 public scene text datasets. The images in the middle two rows are our collected license plate images with text in one-line or multi-line. The last two rows are our synthesized license plate images.}
\label{data}
\vspace{-5mm}
\end{figure}

\textbf{Multi-Line Text 280 (MLT280)} is collected from the Internet and consists of 280 license plate images with text in one-line or two-line. Some samples are shown in Figure~\ref{data}. For each image, we annotate the sequence of characters in order from left to right and top to bottom. No lexicon is provided.

\subsection{Implementation Details}

\noindent\textbf{Network Settings} 
The CNN encoder of our model is adapted from~\cite{bshi2018aster}. Specifically, we only keep the first two $2 \times 2$-stride convolutions for feature maps down-sampling, and the last three $2 \times 2$-stride convolutions in~\cite{bshi2018aster} are replaced with $1 \times 1$-stride convolutions.

To reduce the computation of the first relation attention module, a $1 \times 1$ convolutional layer is applied to the input feature maps to reduce the channel size to 128. By default, the number of transformer layers of relation attention module is set to 2. And for each transformer unit in relation attention module, the hidden size and the number of self-attention heads are set to 512 and 4 respectively. 

As for the parallel attention module, the number of output nodes $n$ is set to 25, since the lengths of most common English words are shorter than 25. Besides, we set the category $|\Omega|$ of the decoded characters to 38,  which corresponds to digits (0-9), English characters (a-z/A-Z, the case is ignored), a symbol ``EOS" to mark the end of a sequence and a symbol ``UNK" to represents all the other characters.

\noindent\textbf{Training} To compare fairly with the previous methods, we train our model on Synth90K and SynthText from scratch. The sample ratio of Synth90K and SynthText is set to $1:1$. In the training stage, the input image is resized to $32 \times 100$. We also use data augmentation (color jittering
, blur \emph{et al.}) as~\cite{DBLP:journals/corr/abs-1809-06508}. Besides, we use ADADELTA~\cite{DBLP:journals/corr/abs-1212-5701} to optimize our model  with the batch size of 128. Empirically, we set the initial learning rate to 1 and decrease it to 0.1 and 0.01 at step 0.6M and 0.8M respectively. And the training phase is terminated at the step 1M.

\noindent\textbf{Inference} In the inference stage, we still resize  the input image to $32\times100$.
We decode the probability matrix to character sequence with the following rules: 1) for each output node, the character with the maximum probability is treated as the output. 2) All ``UNK" symbol will be removed. 3) The decoding process will be terminated when meeting the first ``EOS". By default, we take the predicted sequence from the second stage decoder as our result. Following the previous methods,  we evaluate our method on all datasets with one model by default.

\noindent\textbf{Implementation}
We implement our method with PyTorch~\cite{Pytorch} and conduct all experiments on a regular workstation. By default, we train our model in parallel with two NVIDIA P40 GPU and evaluate on a single GPU with the batch size of 1.

\subsection{Comparisons with State-of-the-Arts}
To verify the effectiveness of our proposed method, we evaluate and compare our method with the previous state-of-the-arts on the above-mentioned benchmarks. Following the previous methods, we measure the recognition performance with word recognition accuracy. An image is considered to be correctly recognized when the predicted text is the same as the ground truth. When a lexicon is available, the predicted text is amended by the word with the minimum edit distance to the predicted text.

\subsubsection{Regular Scene Text Recognition}
We first evaluate our model on 4 regular scene text datasets: IIIT5K, SVT, IC03, IC13. The results are reported in Table~\ref{tab:results}. In the case of fair comparison, our method achieves state-of-the-art results on most datasets. As for IIIT5K and SVT, since some text instances are curved or oriented, our results are slightly better than~\cite{bshi2018aster}, and outperforms the other methods by a large margin. On IC03, our results are comparable to~\cite{cluo2019moran} when evaluated without lexicon, and outperforms all the
other methods when recognizing without lexicon. Our result on IC13 is slightly lower than~\cite{DBLP:conf/iccv/ChengBXZPZ17,DBLP:conf/cvpr/BaiCNPZ18} which is designed for regular text recognition. The performances on regular scene text datasets show the generality and robustness of our proposed method.

\begin{table*}
\centering{}%
\scalebox{0.96}{
\begin{tabular}{|c|c|c|c|c|c|c|c|c|c|c|c|c|}
\hline 
\multirow{2}{*}{Methods} & \multicolumn{3}{c|}{IIIT5K} & \multicolumn{2}{c|}{SVT} & \multicolumn{3}{c|}{IC03} & IC13 & IC15 & SVTP & CUTE\tabularnewline
\cline{2-13} 
 & 50 & 1000 & 0 & 50 & 0 & 50 & Full & 0 & 0 & 0 & 0 & 0\tabularnewline
\hline 
Wang~\emph{et al.}~\cite{DBLP:conf/iccv/WangBB11}  & - & - & - & 57.0 & - & 76.0 & 62.0 & - & - & -& -& - \tabularnewline
Mishra~\emph{et al.}~\cite{DBLP:conf/cvpr/MishraAJ12}  & 64.1 & 57.5 & - & 73.2 & - & 81.8 & 67.8 & - & -& -& - & -\tabularnewline
Wang~\emph{et al.}~\cite{DBLP:conf/icpr/WangWCN12}  & - & - & - & 70.0 & - & 90.0 & 84.0 & - & -& -& - & -\tabularnewline
Bissacco~\emph{et al.}~\cite{DBLP:conf/iccv/BissaccoCNN13}  & - & - & - & - & - & 90.4 & 78.0 & - & 87.6& -& - & -\tabularnewline
Almaz{\'{a}}n~\emph{et al.}~\cite{DBLP:journals/pami/AlmazanGFV14}  & 91.2 & 82.1 & - & 89.2 & - & - & - & - & -& -& - & -\tabularnewline
Yao~\emph{et al.}~\cite{DBLP:conf/cvpr/YaoBSL14}  & 80.2 & 69.3 & - & 75.9 & - & 88.5 & 80.3 & - & -& -& - & -\tabularnewline
Rodr{\'{\i}}guez{-}Serrano~\emph{et al.}~\cite{DBLP:journals/ijcv/Rodriguez-Serrano15}  & 76.1 & 57.4 & - & 70.0 & - & - & - & - & -& -& - & -\tabularnewline
Jaderberg~\emph{et al.}~\cite{DBLP:conf/eccv/JaderbergVZ14}  & - & - & - & 86.1 & - & 96.2 & 91.5 & - & -& -& - & -\tabularnewline
Su and Lu~\cite{DBLP:conf/accv/SuL14}  & - & - & - & 83.0 & - & 92.0 & 82.0 & - & -& -& - & -\tabularnewline
Gordo~\cite{DBLP:conf/cvpr/Gordo15}  & 93.3 & 86.6 & - & 91.8 & - & - & - & - & -& -& - & -\tabularnewline

Jaderberg~\emph{et al.}~\cite{DBLP:journals/corr/JaderbergSVZ14b}  & 95.5 & 89.6 & - & 93.2 & 71.7 & 97.8 & 97.0 & 89.6 & 81.8& -& - & -\tabularnewline
Jaderberg~\emph{et al.}~\cite{DBLP:journals/ijcv/JaderbergSVZ16}  & 97.1 & 92.7 & - & 95.4 & 80.7 & 98.7 & 98.6 & 93.1 & 90.8& -& - & -\tabularnewline

Shi~\emph{et al.}~\cite{DBLP:journals/pami/ShiBY17}  & 97.8 & 95.0 & 81.2 & 97.5 & 82.7 & 98.7 & 98.0 & 91.9 & 89.6& -& - & -\tabularnewline
Shi~\emph{et al.}~\cite{ShiWLYB16}  & 96.2 & 93.8 & 81.9 & 95.5 & 81.9 &98.3 & 96.2 & 90.1& 88.6& - & 71.8 & 59.2\tabularnewline
Lee~\emph{et al.}~\cite{DBLP:conf/cvpr/LeeO16}  & 96.8 & 94.4 & 78.4 & 96.3 & 80.7 & 97.9 & 97.0 & 88.7 & 90.0& -& - & -\tabularnewline
Liu~\emph{et al.}~\cite{DBLP:conf/aaai/LiuCW18} & - & - & 83.6 & - & 84.4 & - & - & 91.5 & 90.8 & - & 73.5 & -\tabularnewline

Yang~\emph{et al.}~\cite{DBLP:conf/ijcai/YangHZKG17}  & 97.8 & 96.1 & - & 95.2 & - & 97.7 & - & - & -& -& 75.8 & 69.3\tabularnewline
Liao~\emph{et al.}~\cite{DBLP:journals/corr/abs-1809-06508}  & 99.8 & 98.8 & 91.9 & 98.8 & 86.4 & - & - & - & 91.5 & -& - & 79.9\tabularnewline
Li~\emph{et al.}~\cite{DBLP:journals/corr/abs-1811-00751}$^\dagger$  & 99.4 & 98.2 & 95.0 & 98.5 & 91.2 & - & - & - & 94.0& 78.8& 86.4 & 89.6\tabularnewline

\hline
\hline
Cheng~\emph{et al.}~\cite{DBLP:conf/iccv/ChengBXZPZ17}*  & 99.3 & 97.5 & 87.4 & 97.1 & 85.9 & 99.2 & 97.3 & 94.2 & 93.3 & 70.6 & - & -\tabularnewline
Cheng~\emph{et al.}~\cite{DBLP:conf/cvpr/ChengXBNPZ18}*  & 99.6 & 98.1 & 87.0 & 96.0 & 82.8 & 98.5 & 97.1 & 91.5 & -& 68.2& 73.0 & 76.8\tabularnewline
Bai~\emph{et al.}~\cite{DBLP:conf/cvpr/BaiCNPZ18}*  & 99.5 & 97.9 & 88.3 & 96.6 & 87.5 & 98.7 & 97.9 & 94.6 & \textbf{94.4}& 73.9& - & -\tabularnewline
Shi~\emph{et al.}~\cite{bshi2018aster}*  & 99.6 & 98.8 & 93.4 & \textbf{97.4} & 89.5 & 98.8 & 98.0 & 94.5 & 91.8& 76.1& 78.5 & 79.5\tabularnewline
Luo~\emph{et al.}~\cite{cluo2019moran}*  & 97.9 & 96.2 & 91.2 & 96.6 & 88.3 & 98.7 & 97.8 & \textbf{95.0} & 92.4& 68.8& 76.1 & 77.4\tabularnewline
Li~\emph{et al.}~\cite{DBLP:journals/corr/abs-1811-00751}*  & - & - & 91.5 & - & 84.5 & - & - & - & 91.0 & 69.2 & 76.4 & 83.3\tabularnewline
Ours & \textbf{99.8} & \textbf{99.1} & \textbf{94.0} & 97.2 & \textbf{90.1} & \textbf{99.4} & \textbf{98.1} & 94.3 & 92.7 & \textbf{76.3} & \textbf{82.3} & \textbf{86.8} \tabularnewline
\hline
\end{tabular}
}
\caption{Results on the public datasets. ``50" and ``1000" mean the size of lexicon. ``Full" represents all the words in the testset are used as lexicon. ``0" means no lexicon is provided. The methods with ``*" indicate that the corresponding model is trained with Synth90k and SynthText, which is the same as ours. So these methods can be compared with ours fairly. Note that the model of~\cite{DBLP:journals/corr/abs-1811-00751}$^\dagger$ is trained on extra synthetic data and some real scene text data. So we just list the results here for reference.}
\label{tab:results}
\vspace{-5mm}
\end{table*}

\vspace{-2mm}
\subsubsection{Irregular Scene Text Recognition}
To validate the effectiveness of our method to recognize irregular scene text, we conduct experiments on three irregular scene text datasets: IC15, SVTP and CUTE. We list and compare our results with previous methods in Table~\ref{tab:results}. Our method beats all the previous methods by a large margin. In detail, our results outperform the previous best results on the three irregular scene text datasets by $0.2\%$, $3.8\%$ and $3.5\%$ respectively. Especially, on CUTE, our method surpasses the rectification-based methods~\cite{bshi2018aster,cluo2019moran} at least $3.5\%$ and performs better
than~\cite{DBLP:conf/cvpr/ChengXBNPZ18} which also recognizes irregular text in 2D perspective by $10\%$. The apparent advantage over previous methods demonstrates that our method is more robust and effective to recognize irregular text, especially in the case of recognizing text with more complex layout (such as the samples in CUTE80).

\vspace{-2mm}
\subsubsection{Multi-line Scene Text Recognition}
Previous irregular scene datasets only contain oriented, perspective or curved text, but lack the text in multi-line. Multi-line text, such as license plate number, mathematical formula and CAPTCHA, is common in natural scene. To verify the capacity of our method to recognize multi-line text, we propose a multi-line scene text dataset, called MLT280. We also create a synthetic license plate dataset which contains 1.2 million synthetic images to train our model. Some  synthetic images are exhibited in Figure~\ref{data}. The MLT 280 and the synthetic images will be made available.

We train two models in total, one trained from scratch with random initialization and the other initialized with the model pre-trained on Synth90K and SynthText, and then fine-tuned on the synthetic license plate dataset. For the model trained from scratch, we train the model with the same training settings as ``Ours" in Table~\ref{tab:results}. As for the fine-tuned model, we train the model about 150K steps, with the learning rate of 0.001. To compare with the previous methods in different pipelines, some other models are trained. ASTER~\cite{bshi2018aster} in rectification-recognition pipeline and SAR~\cite{DBLP:journals/corr/abs-1811-00751} in recurrent attention pipeline are chosen for comparison. We train ASTER and SAR using the official codes in~\cite{aster,sar}. Similarly, two models for each method are trained on the synthetic license plate dataset.

\begin{figure*}
\begin{centering}
\includegraphics[scale=0.9]{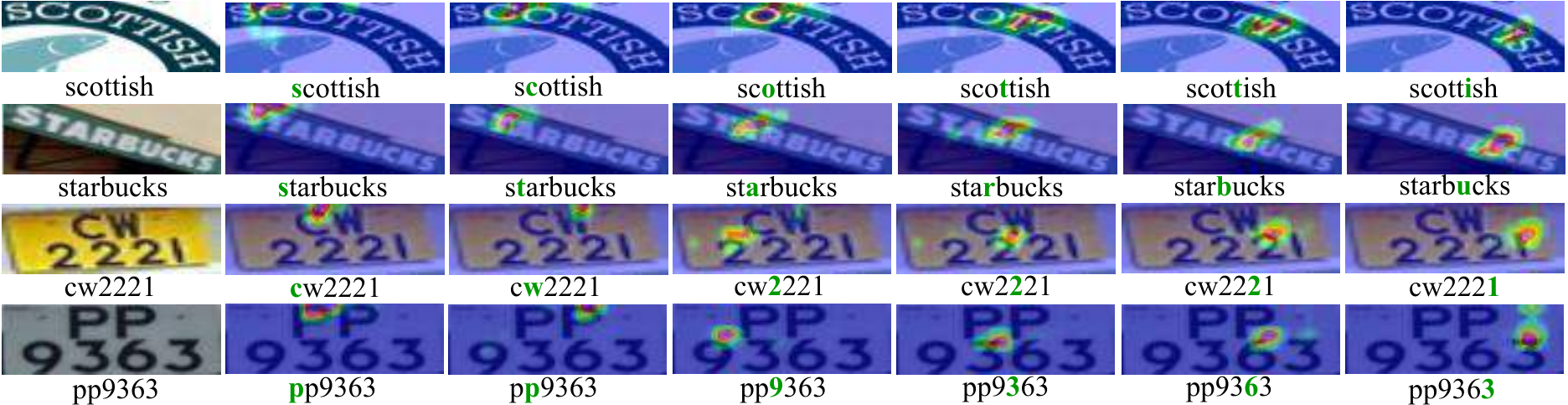}
\par\end{centering}
\caption{Visualization of the 2D attention masks yielded by our parallel attention module. Due to the space limitation, only the first 6 attention masks are presented. More samples are shown in  the supplementary material.}
\label{res_vis}
\vspace{-4mm}
\end{figure*}

The results are summarized in Table~\ref{tab:multi-line}. For ASTER, $40\%$ and $62.5\%$ accuracy are achieved by the random initialized model and fine-tuned model, respectively. We observe that the rectification network of ASTER can not handle text in multi-line since almost all multi-line text images are wrongly recognized, which indicates that the methods in rectification-recognition pipeline are not suitable for multi-line text recognition. As for SAR, $43.9\%$ and $51.1\%$ accuracy are achieved respectively. We visualize some failed cases as well as the corresponding attention masks of SAR in the supplementary material. Those masks show that SAR can not locate the characters accurately. The LSTM encoder in SAR transforms the 2D feature map to 1D sequence by a vertical max-pooling, which may cause inaccurate localization due to the loss of 2D space information.  Our method achieves the best results in both cases.  Particularly, $61.4\%$ and $80.7\%$ accuracy are achieved by the random initialized model and fine-tuned model, which outperform ASTER and SAR in a large margin and demonstrate that our method can handle the multi-line text recognition task well. Some qualitative results are presented in Figure~\ref{res_vis}. The accurate localization in the attention masks indicates that the multi-line text can be  well handled by our method.

\begin{table}
\begin{centering}
\scalebox{0.87}{
\begin{tabular}{|c|c|c|c|c|}
\hline 
\multirow{2}{*}{Methods} & \multicolumn{2}{c|}{Accuracy} & \multicolumn{2}{c|}{Speed}\tabularnewline
\cline{2-5} 
 & random init & fine-tuned & P40 & Original\tabularnewline
\hline 
ASTER~\cite{bshi2018aster} & 40.0 & 62.5 & 32.4ms & 20ms \tabularnewline
\hline 
SAR~\cite{DBLP:journals/corr/abs-1811-00751} & 43.9  & 51.1  & 66.7ms  & 15ms* \tabularnewline
\hline 
Ours & \textbf{61.4} & \textbf{80.7} & \textbf{15.1ms} & - \tabularnewline
\hline 
\end{tabular}
}
\par\end{centering}
\caption{The comparison of accuracy and speed between other irregular scene text recognition methods and ours on MLT280. ``Original" means the speed reported in the original paper which is evaluated on different hardware platforms. *The original result is tested on NVIDIA Titan x GPU, and the batch size is unclear.}
\label{tab:multi-line}
\vspace{-4mm}
\end{table}

\subsubsection{Speed}
To prove the efficiency of our proposed method, we evaluate the speed of our method as well as other irregular scene text recognizers. To compare fairly, we evaluate all methods on the same hardware platform and set the test batch size to 1. Each method is tested 5 times on MLT280. And the average speed for a single image is listed in Table~\ref{tab:multi-line}. Benefiting from the parallel computing of our proposed relation attention module and parallel attention module, our model is \textbf{2.1} times faster than the rectification-based method~\cite{bshi2018aster} and \textbf{4.4} times faster than the recurrent 2D attention based method~\cite{DBLP:journals/corr/abs-1811-00751}.

\subsection{Ablation Study}

\subsubsection{The Effectiveness of the Second Stage Decoder}
To prove the effectiveness of the second stage decoder, we compare the results yielded by the first stage decoder and the second stage decoder quantitatively. For convenience, the results of the first decoder and the second decoder are named ``Ours(d1)" and ``Ours(d2)" and listed in Table~\ref{tab:d1d2}. As shown, ``Ours(d2)" is always better than ``Ours(d1)" on all datasets. It indicates that, to a certain extent the dependency relationships among the glimpses can be captured by the relation attention module of the second stage.
\begin{table}
\begin{centering}
\scalebox{0.8}{
    \begin{tabular}{|c|c|c|c|c|c|c|c|}
    \hline 
    Methods & IIIT5K & SVT & IC03 & IC13 & IC15 & SVTP & CUTE\tabularnewline
    \hline 
    Ours(d1) & 93.3  & 89.3  & 94 & 92.5 & 75.4 & 81.9 & 86.5 \tabularnewline
    \hline 
    Ours(d2) & \textbf{94.0} & \textbf{90.1} & \textbf{94.3} & \textbf{92.7} & \textbf{76.3} & \textbf{82.3}  & \textbf{86.8} \tabularnewline
    \hline 
    \end{tabular}
}
\par\end{centering}
\caption{The comparison of the first decoder and the second decoder.}
\label{tab:d1d2}
\vspace{-5mm}
\end{table}
\vspace{-2mm}
\subsubsection{The Effectiveness of Relation Attention Module}
 We also evaluate our model with different number of transformer layers in a relation attention module to assess the effectiveness of relation attention module. The results are presented in Table~\ref{tab:relation}. For convenience, we name our models as ``Ours(a,b)", where $a$ and $b$ are the number of transformer layers of the first and the second relation attention module respectively. The value will be set to $0$ if the corresponding relation attention module is not used. As shown in Table~\ref{tab:relation}, the models with relation attention modules are always better than the one without. The comparison of ``Ours(0, 0)" and ``Ours(2, 0)" shows the first relation attention module can improve the recognition accuracy both on regular and irregular scene text datasets. The comparison of ``Ours(2, 0)" and ``Ours(2, 2)" demonstrates that the second relation attention module which captures the dependency relationships of output nodes can further improve the accuracy. We also evaluate our models with more transformer layers in relation attention module. Comparable results with ``Ours(2,2)" are achieved. For the trade-off of speed and accuracy, we use the setting (2, 2) by default.

\begin{table}
\begin{centering}
\scalebox{0.8}{
    \begin{tabular}{|c|c|c|c|c|c|c|c|}
    \hline 
    Methods & IIIT5K & SVT & IC03 & IC13 & IC15 & SVTP & CUTE\tabularnewline
    \hline 
    Ours(0,0) & 92.1 & 88.1 & 94.1 & 91.5 & 73.4 & 80 & 85.4 \tabularnewline
    \hline 
    Ours(2,0) & 93.5 & 90.3 & \textbf{94.3} & 92.2 & 74 & 80.9 & 85.1\tabularnewline
    \hline 
    Ours(2,2) & \textbf{94.0} & \textbf{90.1} & \textbf{94.3} & \textbf{92.7} & \textbf{76.3} & \textbf{82.3}  & \textbf{86.8} \tabularnewline
    \hline 
    \end{tabular}
}
\par\end{centering}
\caption{The effect of relation attention module.}
\label{tab:relation}
\end{table}



\subsection{Limitation}
\begin{figure}
\begin{centering}
\includegraphics[scale=0.82]{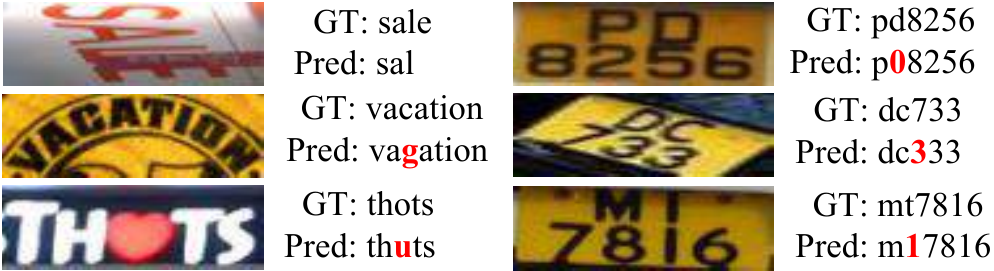}
\par\end{centering}
\caption{Some failed cases of our method. The wrongly recognized characters are marked in red.}
\label{failure cases}
\vspace{-4mm}
\end{figure}

We visualize some failure cases of our method in Figure~\ref{failure cases}. As shown, our method fails to recognize the vertical text images, since there are no vertical text images in training samples. Besides our method also struggles to recognize the characters with the similar appearance (such as ``G" and ``C", ``D" and ``0") as well as the images with complex environmental interference (occlusion, blur~\emph{et al.}).

\section{Conclusion}
In this paper, we present a new irregular scene text recognition method. The proposed method which recognizes text image with 2D image feature can reserve and utilize the 2D space information of text directly. Besides, benefiting from the efficiency of the proposed relation attention module and parallel attention module, our method is faster than the previous methods. We evaluate our model both on 7 public datasets as well as our proposed multi-line text dataset. The superior performances on all datasets prove the effectiveness and efficiency of our proposed method. In the future, recognizing text in  vertical and more complex layouts will be a goal for us. 
\appendix
\section{Appendix}
\subsection{Architecture of Relation Attention Module}
\begin{figure*}
\begin{centering}
\includegraphics[scale=0.6]{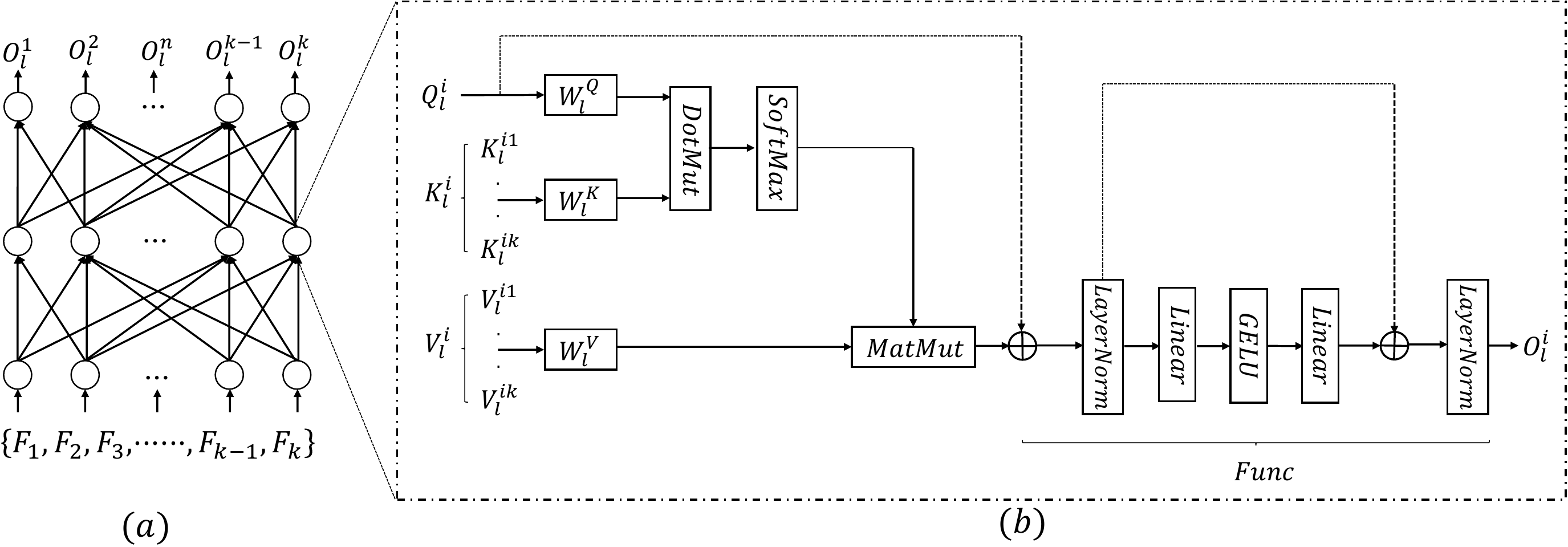}
\par\end{centering}
\caption{Architecture of our proposed Relation Attention Module. a) Overview of the relation attention module; b) The details of a transformer unit. }
\label{relation attention}
\end{figure*}

The architecture of our proposed relation attention module is shown in Figure~\ref{relation attention}(a). As shown, the relation attention module is in multi-layer architecture and consists of transformer units , one of which is presented in Figure~\ref{relation attention}(b). For the first layer, the inputs are the fused features $F$, which are the sum of the input features $I$ and the position embedding features $E$. For the others layers, the inputs are the outputs of the previous layer. In detail, the query, keys and values of each transformer unit are obtained by the Eq.1, Eq.2 and Eq.3 respectively.

\subsection{Qualitative Analysis}
\begin{figure*}
\begin{centering}
\includegraphics[scale=0.95]{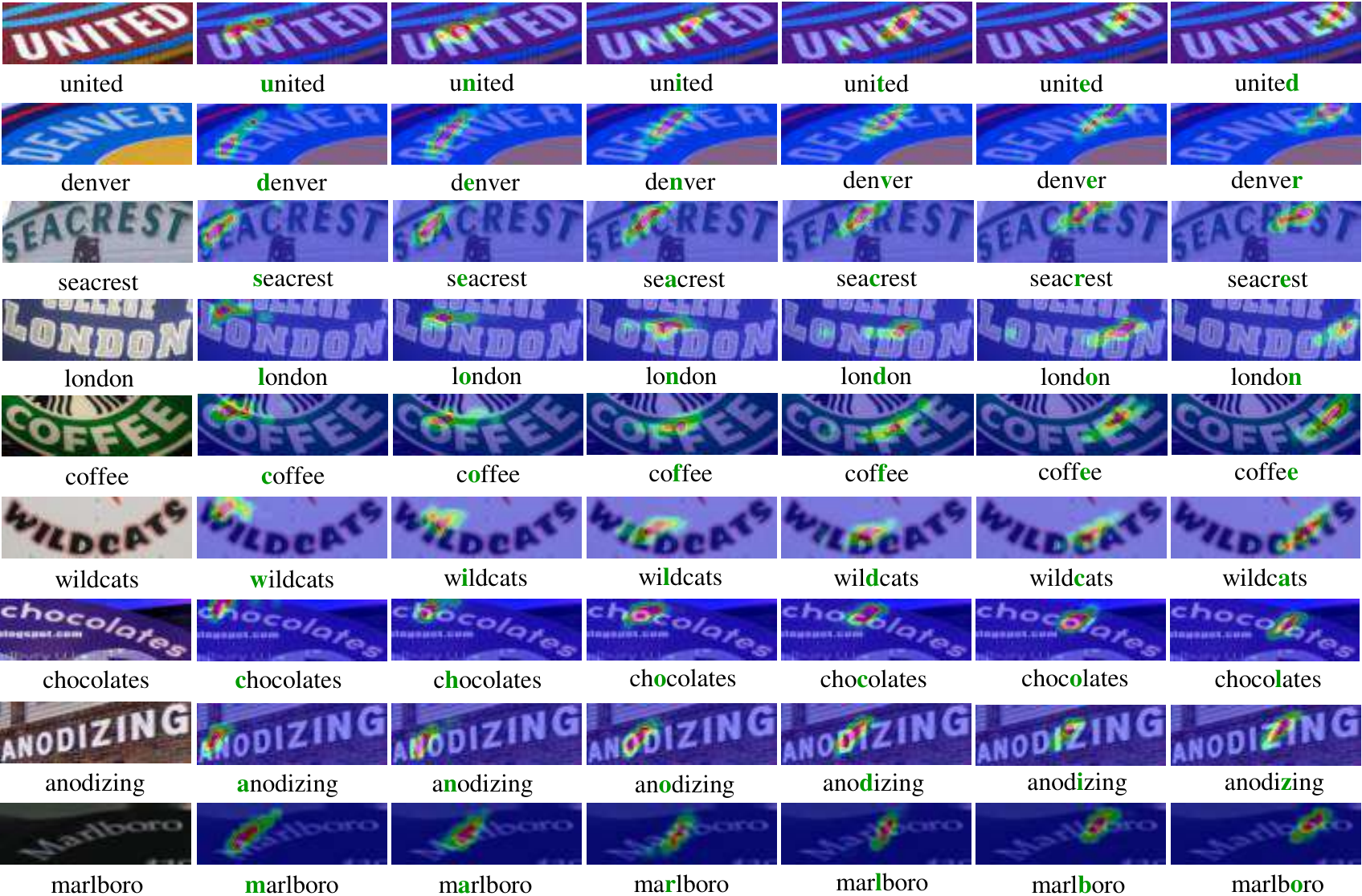}
\par\end{centering}
\caption{Visualization of 2D attention masks yielded by our parallel attention module.}
\label{heatmap}
\end{figure*}

We show more qualitative results of our method in Figure~\ref{heatmap}. As shown, our method is effective to cope with irregular scene text such as curved text images and oblique text images. The highlighted text area of 2D attention masks indicates that our method is able to learn the positions of characters and recognize irregular scene text.

\subsection{Qualitative Comparison}
\begin{figure*}[htbp]
\begin{centering}
\includegraphics[scale=0.95]{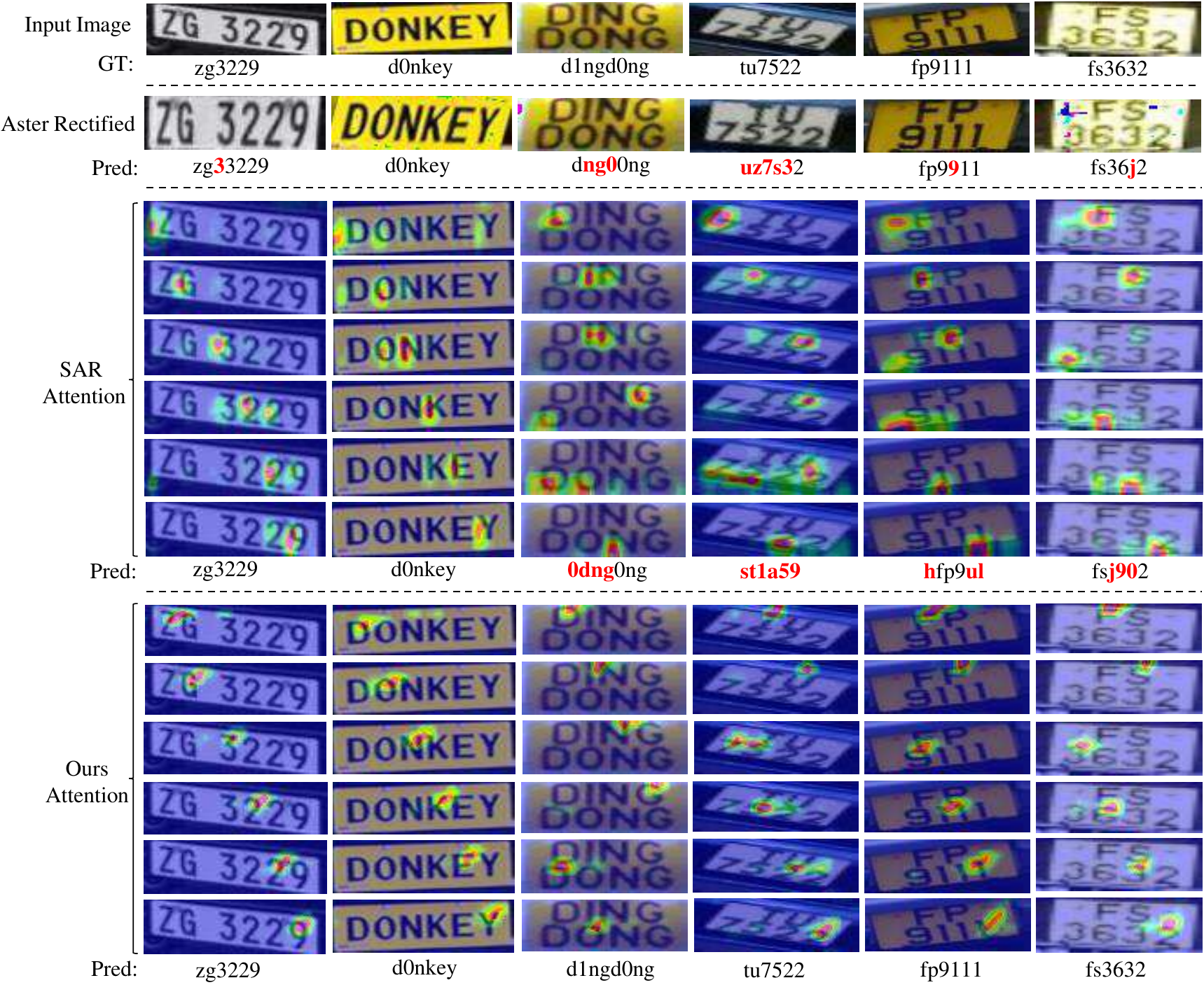}
\par\end{centering}
\caption{Comparison of our method with other competitors on MLT280. Note that, to avoid ambiguity with ``1'' and ``0", the letters "I", "O" and "Q" are not included in the character set.}
\label{heatmap compare}
\end{figure*}

This section describes the qualitative comparison of our method with ASTER~\cite{bshi2018aster} and SAR~\cite{DBLP:journals/corr/abs-1811-00751}. Some samples from our collected dataset MLT280 are presented in Figure~\ref{heatmap compare}. Specifically, the first part is the input image and corresponding ground truth; the second part is the rectified image generated by rectification network of ASTER and the corresponding prediction; the third and fourth part are the attention masks and predictions of SAR and ours respectively. The results show that multi-line text images can not be well handled by ASTER, which may be due to the limited capacity of the rectification network. The visualization of 2D attention masks of SAR shows that SAR can learn the position information of one-line text well, but can not address complex multi-line text scenario as well. This phenomenon may be caused by the irreversible 2D spacial information loss of vertical pooling. Since we reserve and utilize 2D spacial information to attend text area, our method can localize and recognize both one-line and multi-line text image.   

{\small
\bibliographystyle{ieee}
\bibliography{egbib}

\begin{thebibliography}{10}\itemsep=-1pt

\bibitem{aster}
Aster.
\newblock \url{https://github.com/bgshih/aster}.

\bibitem{Pytorch}
Pytorch.
\newblock \url{https://pytorch.org/}.

\bibitem{sar}
Sar.
\newblock
  \url{https://github.com/wangpengnorman/SAR-Strong-Baseline-for-Text-Recognition}.

\bibitem{Synth90K}
Synth90k.
\newblock \url{http://www.robots.ox.ac.uk/~vgg/data/text/}.

\bibitem{SynthText}
Synthtext.
\newblock \url{http://www.robots.ox.ac.uk/~vgg/data/scenetext/}.

\bibitem{DBLP:journals/pami/AlmazanGFV14}
J.~Almaz{\'{a}}n, A.~Gordo, A.~Forn{\'{e}}s, and E.~Valveny.
\newblock Word spotting and recognition with embedded attributes.
\newblock {\em {IEEE} Trans. Pattern Anal. Mach. Intell.}, 2014.

\bibitem{DBLP:journals/corr/BahdanauCB14}
D.~Bahdanau, K.~Cho, and Y.~Bengio.
\newblock Neural machine translation by jointly learning to align and
  translate.
\newblock In {\em ICLR}, 2015.

\bibitem{DBLP:conf/cvpr/BaiCNPZ18}
F.~Bai, Z.~Cheng, Y.~Niu, S.~Pu, and S.~Zhou.
\newblock Edit probability for scene text recognition.
\newblock In {\em CVPR}, 2018.

\bibitem{DBLP:conf/iccv/BissaccoCNN13}
A.~Bissacco, M.~Cummins, Y.~Netzer, and H.~Neven.
\newblock Photoocr: Reading text in uncontrolled conditions.
\newblock In {\em ICCV}, 2013.

\bibitem{DBLP:conf/iccv/BustaNM17}
M.~Busta, L.~Neumann, and J.~Matas.
\newblock Deep textspotter: An end-to-end trainable scene text localization and
  recognition framework.
\newblock In {\em ICCV}, 2017.

\bibitem{DBLP:conf/iccv/ChengBXZPZ17}
Z.~Cheng, F.~Bai, Y.~Xu, G.~Zheng, S.~Pu, and S.~Zhou.
\newblock Focusing attention: Towards accurate text recognition in natural
  images.
\newblock In {\em ICCV}, 2017.

\bibitem{DBLP:conf/cvpr/ChengXBNPZ18}
Z.~Cheng, Y.~Xu, F.~Bai, Y.~Niu, S.~Pu, and S.~Zhou.
\newblock {AON:} towards arbitrarily-oriented text recognition.
\newblock In {\em CVPR}, 2018.

\bibitem{DBLP:conf/nips/ChorowskiBSCB15}
J.~Chorowski, D.~Bahdanau, D.~Serdyuk, K.~Cho, and Y.~Bengio.
\newblock Attention-based models for speech recognition.
\newblock In {\em NIPS}, 2015.

\bibitem{DBLP:journals/corr/abs-1810-04805}
J.~Devlin, M.~Chang, K.~Lee, and K.~Toutanova.
\newblock {BERT:} pre-training of deep bidirectional transformers for language
  understanding.
\newblock {\em arXiv preprint arXiv:1810.04805}, 2018.

\bibitem{DBLP:conf/cvpr/Gordo15}
A.~Gordo.
\newblock Supervised mid-level features for word image representation.
\newblock In {\em CVPR}, 2015.

\bibitem{DBLP:conf/icml/GravesFGS06}
A.~Graves, S.~Fern{\'{a}}ndez, F.~J. Gomez, and J.~Schmidhuber.
\newblock Connectionist temporal classification: labelling unsegmented sequence
  data with recurrent neural networks.
\newblock In {\em ICML}, 2006.

\bibitem{DBLP:conf/cvpr/GuptaVZ16}
A.~Gupta, A.~Vedaldi, and A.~Zisserman.
\newblock Synthetic data for text localisation in natural images.
\newblock In {\em CVPR}, 2016.

\bibitem{DBLP:conf/cvpr/HeTHS0S18}
T.~He, Z.~Tian, W.~Huang, C.~Shen, Y.~Qiao, and C.~Sun.
\newblock An end-to-end textspotter with explicit alignment and attention.
\newblock In {\em CVPR}, 2018.

\bibitem{DBLP:journals/neco/HochreiterS97}
S.~Hochreiter and J.~Schmidhuber.
\newblock Long short-term memory.
\newblock {\em Neural Computation}, 1997.

\bibitem{DBLP:conf/cvpr/HuGZDW18}
H.~Hu, J.~Gu, Z.~Zhang, J.~Dai, and Y.~Wei.
\newblock Relation networks for object detection.
\newblock In {\em CVPR}, 2018.

\bibitem{DBLP:journals/corr/JaderbergSVZ14b}
M.~Jaderberg, K.~Simonyan, A.~Vedaldi, and A.~Zisserman.
\newblock Deep structured output learning for unconstrained text recognition.
\newblock {\em arXiv preprint arXiv:1412.5903}, 2014.

\bibitem{DBLP:journals/corr/JaderbergSVZ14}
M.~Jaderberg, K.~Simonyan, A.~Vedaldi, and A.~Zisserman.
\newblock Synthetic data and artificial neural networks for natural scene text
  recognition.
\newblock {\em arXiv preprint arXiv:1406.2227}, 2014.

\bibitem{DBLP:journals/ijcv/JaderbergSVZ16}
M.~Jaderberg, K.~Simonyan, A.~Vedaldi, and A.~Zisserman.
\newblock Reading text in the wild with convolutional neural networks.
\newblock {\em International Journal of Computer Vision}, 2016.

\bibitem{DBLP:conf/nips/JaderbergSZK15}
M.~Jaderberg, K.~Simonyan, A.~Zisserman, and K.~Kavukcuoglu.
\newblock Spatial transformer networks.
\newblock In {\em NIPS}, 2015.

\bibitem{DBLP:conf/eccv/JaderbergVZ14}
M.~Jaderberg, A.~Vedaldi, and A.~Zisserman.
\newblock Deep features for text spotting.
\newblock In {\em ECCV}, 2014.

\bibitem{DBLP:conf/icdar/KaratzasGNGBIMN15}
D.~Karatzas, L.~Gomez{-}Bigorda, A.~Nicolaou, S.~K. Ghosh, A.~D. Bagdanov,
  M.~Iwamura, J.~Matas, L.~Neumann, V.~R. Chandrasekhar, S.~Lu, F.~Shafait,
  S.~Uchida, and E.~Valveny.
\newblock {ICDAR} 2015 competition on robust reading.
\newblock In {\em ICDAR}, 2015.

\bibitem{DBLP:conf/icdar/KaratzasSUIBMMMAH13}
D.~Karatzas, F.~Shafait, S.~Uchida, M.~Iwamura, L.~G. i~Bigorda, S.~R. Mestre,
  J.~Mas, D.~F. Mota, J.~Almaz{\'{a}}n, and L.~de~las Heras.
\newblock {ICDAR} 2013 robust reading competition.
\newblock In {\em ICDAR}, 2013.

\bibitem{lecun1998gradient}
Y.~LeCun, L.~Bottou, Y.~Bengio, P.~Haffner, et~al.
\newblock Gradient-based learning applied to document recognition.
\newblock {\em Proceedings of the IEEE}, 1998.

\bibitem{DBLP:conf/cvpr/LeeBDJP14}
C.~Lee, A.~Bhardwaj, W.~Di, V.~Jagadeesh, and R.~Piramuthu.
\newblock Region-based discriminative feature pooling for scene text
  recognition.
\newblock In {\em CVPR}, 2014.

\bibitem{DBLP:conf/cvpr/LeeO16}
C.~Lee and S.~Osindero.
\newblock Recursive recurrent nets with attention modeling for {OCR} in the
  wild.
\newblock In {\em CVPR}, 2016.

\bibitem{DBLP:journals/corr/abs-1811-00751}
H.~Li, P.~Wang, C.~Shen, and G.~Zhang.
\newblock Show, attend and read: {A} simple and strong baseline for irregular
  text recognition.
\newblock In {\em AAAI}, 2019.

\bibitem{DBLP:journals/corr/abs-1809-06508}
M.~Liao, J.~Zhang, Z.~Wan, F.~Xie, J.~Liang, P.~Lyu, C.~Yao, and X.~Bai.
\newblock Scene text recognition from two-dimensional perspective.
\newblock In {\em AAAI}, 2018.

\bibitem{DBLP:conf/aaai/LiuCW18}
W.~Liu, C.~Chen, and K.~K. Wong.
\newblock Char-net: {A} character-aware neural network for distorted scene text
  recognition.
\newblock In {\em AAAI}, 2018.

\bibitem{DBLP:conf/cvpr/LiuLYCQY18}
X.~Liu, D.~Liang, S.~Yan, D.~Chen, Y.~Qiao, and J.~Yan.
\newblock {FOTS:} fast oriented text spotting with a unified network.
\newblock In {\em CVPR}, 2018.

\bibitem{DBLP:journals/ijdar/LucasPSTWYANOYMZOWJTWL05}
S.~M. Lucas, A.~Panaretos, L.~Sosa, A.~Tang, S.~Wong, R.~Young, K.~Ashida,
  H.~Nagai, M.~Okamoto, H.~Yamamoto, H.~Miyao, J.~Zhu, W.~Ou, C.~Wolf,
  J.~Jolion, L.~Todoran, M.~Worring, and X.~Lin.
\newblock {ICDAR} 2003 robust reading competitions: entries, results, and
  future directions.
\newblock {\em {IJDAR}}, 2005.

\bibitem{cluo2019moran}
C.~Luo, L.~Jin, and Z.~Sun.
\newblock Moran: A multi-object rectified attention network for scene text
  recognition.
\newblock {\em Pattern Recognition}, 2019.

\bibitem{DBLP:conf/eccv/LyuLYWB18}
P.~Lyu, M.~Liao, C.~Yao, W.~Wu, and X.~Bai.
\newblock Mask textspotter: An end-to-end trainable neural network for spotting
  text with arbitrary shapes.
\newblock In {\em ECCV}, 2018.

\bibitem{DBLP:conf/bmvc/MishraAJ12}
A.~Mishra, K.~Alahari, and C.~V. Jawahar.
\newblock Scene text recognition using higher order language priors.
\newblock In {\em BMVC}, 2012.

\bibitem{DBLP:conf/cvpr/MishraAJ12}
A.~Mishra, K.~Alahari, and C.~V. Jawahar.
\newblock Top-down and bottom-up cues for scene text recognition.
\newblock In {\em CVPR}, 2012.

\bibitem{DBLP:conf/cvpr/NeumannM12}
L.~Neumann and J.~Matas.
\newblock Real-time scene text localization and recognition.
\newblock In {\em CVPR}, 2012.

\bibitem{DBLP:conf/iccv/PhanSTT13}
T.~Q. Phan, P.~Shivakumara, S.~Tian, and C.~L. Tan.
\newblock Recognizing text with perspective distortion in natural scenes.
\newblock In {\em ICCV}, 2013.

\bibitem{DBLP:journals/eswa/RisnumawanSCT14}
A.~Risnumawan, P.~Shivakumara, C.~S. Chan, and C.~L. Tan.
\newblock A robust arbitrary text detection system for natural scene images.
\newblock {\em Expert Syst. Appl.}, 2014.

\bibitem{DBLP:journals/ijcv/Rodriguez-Serrano15}
J.~A. Rodr{\'{\i}}guez{-}Serrano, A.~Gordo, and F.~Perronnin.
\newblock Label embedding: {A} frugal baseline for text recognition.
\newblock {\em International Journal of Computer Vision}, 2015.

\bibitem{DBLP:journals/pami/ShiBY17}
B.~Shi, X.~Bai, and C.~Yao.
\newblock An end-to-end trainable neural network for image-based sequence
  recognition and its application to scene text recognition.
\newblock {\em {IEEE} Trans. Pattern Anal. Mach. Intell.}, 2017.

\bibitem{ShiWLYB16}
B.~Shi, X.~Wang, P.~Lyu, C.~Yao, and X.~Bai.
\newblock Robust scene text recognition with automatic rectification.
\newblock In {\em CVPR}, 2016.

\bibitem{bshi2018aster}
B.~Shi, M.~Yang, X.~Wang, P.~Lyu, C.~Yao, and X.~Bai.
\newblock Aster: An attentional scene text recognizer with flexible
  rectification.
\newblock {\em IEEE Transactions on Pattern Analysis and Machine Intelligence},
  2018.

\bibitem{DBLP:conf/accv/SuL14}
B.~Su and S.~Lu.
\newblock Accurate scene text recognition based on recurrent neural network.
\newblock In {\em ACCV}, 2014.

\bibitem{DBLP:conf/nips/VaswaniSPUJGKP17}
A.~Vaswani, N.~Shazeer, N.~Parmar, J.~Uszkoreit, L.~Jones, A.~N. Gomez,
  L.~Kaiser, and I.~Polosukhin.
\newblock Attention is all you need.
\newblock In {\em NIPS}, 2017.

\bibitem{DBLP:conf/iccv/WangBB11}
K.~Wang, B.~Babenko, and S.~J. Belongie.
\newblock End-to-end scene text recognition.
\newblock In {\em ICCV}, 2011.

\bibitem{DBLP:conf/icpr/WangWCN12}
T.~Wang, D.~J. Wu, A.~Coates, and A.~Y. Ng.
\newblock End-to-end text recognition with convolutional neural networks.
\newblock In {\em ICPR}, 2012.

\bibitem{DBLP:conf/ijcai/YangHZKG17}
X.~Yang, D.~He, Z.~Zhou, D.~Kifer, and C.~L. Giles.
\newblock Learning to read irregular text with attention mechanisms.
\newblock In {\em IJCAI}, 2017.

\bibitem{DBLP:journals/tip/YaoBL14}
C.~Yao, X.~Bai, and W.~Liu.
\newblock A unified framework for multioriented text detection and recognition.
\newblock {\em {IEEE} Trans. Image Processing}, 2014.

\bibitem{DBLP:conf/cvpr/YaoBSL14}
C.~Yao, X.~Bai, B.~Shi, and W.~Liu.
\newblock Strokelets: {A} learned multi-scale representation for scene text
  recognition.
\newblock In {\em CVPR}, 2014.

\bibitem{DBLP:journals/corr/abs-1212-5701}
M.~D. Zeiler.
\newblock {ADADELTA:} an adaptive learning rate method.
\newblock {\em arXiv preprint arXiv:1212.5701}, 2012.

\end{thebibliography}
}

\end{document}